\documentclass[runningheads]{llncs}
\usepackage[T1]{fontenc}
\usepackage{graphicx,verbatim}
\usepackage{amsmath}
\usepackage{multirow}
\usepackage{algorithm}
\usepackage{algpseudocode}
\usepackage{subcaption}
\usepackage{graphicx}
\usepackage{caption}
\usepackage{booktabs}

\begin{document}
\title{Semi-supervised Liver Segmentation and Patch-based Fibrosis Staging with Registration-aided Multi-parametric MRI}

\titlerunning{Liver Segmentation and Fibrosis Staging} 

\author{Boya Wang\inst{1}\thanks{Corresponding author: boya.wang@nottingham.ac.uk}, Ruizhe Li\inst{1, 2, 3}, Chao Chen\inst{1, 2} and Xin Chen\inst{1}}  
\authorrunning{Boya Wang et al.}

\institute{
Intelligent Modelling \& Analysis Group (IMA), School of Computer Science, University of Nottingham, UK \and
Lab for Uncertainty in Data and Decision Making (LUCID), School of Computer Science, University of Nottingham, UK \and
Nottingham Biomedical Research Centre (BRC), School of Medicine, University of Nottingham, UK
}
\maketitle  

\begin{abstract}
Liver fibrosis poses a substantial challenge in clinical practice, emphasizing the necessity for precise liver segmentation and accurate disease staging. Based on the CARE Liver 2025 Track 4 Challenge, this study introduces a multi-task deep learning framework developed for liver segmentation (LiSeg) and liver fibrosis staging (LiFS) using multi-parametric MRI. The LiSeg phase addresses the challenge of limited annotated images and the complexities of multi-parametric MRI data by employing a semi-supervised learning model that integrates image segmentation and registration. By leveraging both labeled and unlabeled data, the model overcomes the difficulties introduced by domain shifts and variations across modalities. In the LiFS phase, we employed a patch-based method which allows the visualization of liver fibrosis stages based on the classification outputs. Our approach effectively handles multi-modality imaging data, limited labels, and domain shifts. The proposed method has been tested by the challenge organizer on an independent test set that includes in-distribution (ID) and out-of-distribution (OOD) cases using three-channel MRIs (T1, T2, DWI) and seven-channel MRIs (T1, T2, DWI, GED1-GED4). The code is freely available. Github link: https://github.com/mileywang3061/Care-Liver


\end{abstract}
\section{Introduction}
\label{sec:intro}

Liver fibrosis can progress to cirrhosis or hepatocellular carcinoma if not detected early. The CARE Liver 2025 challenge (Track 4)~\cite{liu2025merit,gao2023reliable,wu2022meru} aims to develop AI-based methods for liver segmentation and fibrosis staging, using multi-center, multi-vendor MRI data from 610 patients with liver fibrosis. The dataset includes T1-weighted (T1), T2-weighted (T2), diffusion-weighted (DWI), and Gd-EOB-DTPA-enhanced dynamic phases (GED1–GED4), covering arterial, portal venous, delayed, and hepatobiliary phases. All data include GED4, which is always available, while other phases may be missing and are not spatially aligned.

This study addresses both segmentation and classification tasks, which are interconnected but require different modeling strategies. For the segmentation task, we focus on a semi-supervised setting, where only a small subset of GED4 images have annotated liver masks, while all other phases remain unlabeled. The multi-vendor and multi-phase nature of the dataset poses challenges such as domain shifts and misalignment between modalities. Previously developed supervised methods perform well on fully labeled, single-modality datasets but struggle with limited annotations and large modality variations. Semi-supervised approaches (e.g., Mean Teacher~\cite{lou2024semi}, pseudo-labeling~\cite{deng2025correlation,jia2022semi}) leverage unlabeled data, but often fail to ensure spatial and structural consistency across modalities. In contrast, joint registration–segmentation frameworks~\cite{xu2019deepatlas,li2025unified,he2022learning} have shown promise by maintaining geometric structure and label consistency through spatial alignment.

Recent advancements in liver fibrosis classification have increasingly leveraged deep learning and radiomics to improve diagnostic accuracy and clinical interpretability. Convolutional neural networks (CNNs) applied to gadoxetic acid-enhanced MR images have demonstrated promising performance~\cite{yasaka2018deep}. More recently, transformer-based architectures, such as the Vision Transformer (ViT), have exploited self-attention mechanisms to enhance feature representation~\cite{dai2021transmed,manzari2023medvit}. Complementary to deep learning, radiomics extracts high-dimensional quantitative features from medical images, including texture, shape, and intensity~\cite{park2019radiomics,wang2019deep}. Current trends emphasize the integration of multi-view learning and uncertainty modeling to simultaneously improve accuracy and interpretability. For example, Gao et al.~\cite{gao2023reliable} proposed a framework combining local and global features to achieve reliable and interpretable results, while MERIT~\cite{liu2025merit} further incorporates feature-specific evidential fusion and class-distribution-aware calibration, outperforming traditional CNN and radiomics-based approaches.

In this study, we propose a framework for liver segmentation and fibrosis staging from multi-phase and multi-vendor liver MRI data. For segmentation, the BRBS framework~\cite{he2022learning} is adopted and improved to jointly learn registration and segmentation in a semi-supervised manner, enabling label propagation across modalities and improving performance with limited annotations. For classification, a patch-based strategy is introduced to capture fine-grained intensity and texture features, where Stage 1 and Stage 4 image patches serve as the surrogates for healthy and unhealthy tissues, and subject-level stages are inferred from the proportion of Stage 4 patches. By integrating both tasks, the framework ensures that segmentation outputs support classification, enhancing reliability and clinical applicability.

\section{Methods}
\label{sec:method}

\begin{figure*}[t]
    \centering
    \includegraphics[width=0.95\linewidth]{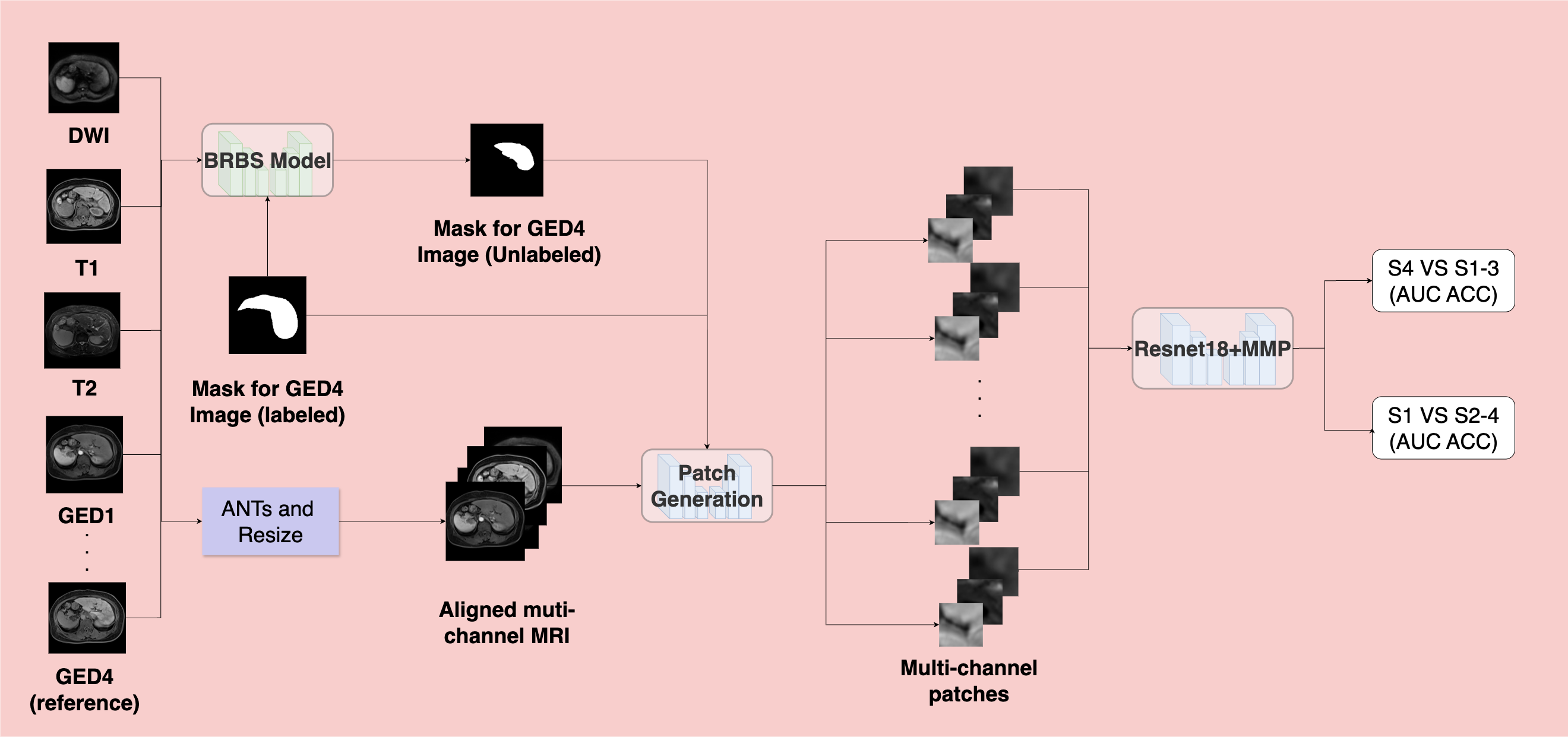}
    \caption{Overview of the proposed framework. Multi-parametric MRI (DWI, T1, T2, GED1–GED4 with GED4 as reference) are aligned using ANTs \cite{tustison2021antsx}, and the liver masks of unlabeled GED4 are generated by the BRBS model. Multi-channel patches are extracted by ResNet18 and probability mapping for cirrhosis (S4 vs.~S1--3) and substantial (S1 vs.~S2--4) fibrosis classification.}
    \label{fig:care2025-seg}
\end{figure*}

As illustrated in Fig.~\ref{fig:care2025-seg}, the overall framework consists of two main stages: semi-supervised multi-parametric MRI segmentation and patch-based fibrosis classification. The multi-parametric MRI with the labeled GED4 masks are first input into the Better Registration Better Segmentation (BRBS) framework \cite{he2022learning} to generate the masks for unlabeled GED4 images. The multi-parametric MRI is also input to the ANTs algorithm \cite{tustison2021antsx} to be spatially aligned. Subsequently, the aligned MRIs and their corresponding GED4 masks are used in the patch-based classification model based on the ResNet18 architecture to achieve fibrosis classification. 



\subsection{Semi-supervised Liver Segmentation and MRI Alignment}
\label{ssec:seg}
The liver segmentation of GED4 images is performed using the BRBS framework \cite{he2022learning}, which jointly learns registration and segmentation in a semi-supervised setting. In the CARE Liver 2025 dataset, only the hepatobiliary phase (GED4) provides manual annotations, while other phases remain unlabeled. BRBS addresses this by registering each unlabeled modality to GED4 and warping the manual mask to generate pseudo-labels, thereby enabling the segmentation network to be supervised across all phases.  

To improve pseudo-label reliability, BRBS introduces two mechanisms. The \textbf{Weighted Consistency Constraint (WCC)} requires the warped atlas mask to match the segmentation prediction, while the estimated deformation fields are constrained to be smooth and cycle-consistent. The \textbf{Space--Style Sampling Program (S3P)} augments supervision by interpolating spatial deformations and style displacements between atlas and another unlabeled image. The resulting synthetic pairs capture intermediate anatomical shapes and contrast variations. Together, WCC and S3P enhance the authenticity, diversity, and robustness of pseudo-labels, enabling reliable semi-supervised liver segmentation across all MRIs.


In the original BRBS design, the registration sub-network is trained using Normalized Cross-Correlation (NCC) as the similarity loss. However, due to the multi-modal nature of the CARE Liver dataset, we replace NCC with a local, patch-based Mutual Information (MI) loss \cite{guo2019multi}, which is more robust to modality-dependent intensity variations.

For each patch $p$, we compute the mutual information between the fixed image $X$ and the moving image $Y$ as:
\begin{equation}
\text{MI}^{(p)}(X, Y) = \sum_{i,j} P_{XY}^{(p)}(i,j)
\log \left( \frac{P_{XY}^{(p)}(i,j) + \varepsilon}{P_X^{(p)}(i) P_Y^{(p)}(j) + \varepsilon} \right)
\end{equation}
where $P_{XY}^{(p)}(i,j)$ is the joint histogram over intensity bins $i$ and $j$ in patch $p$, and $P_X^{(p)}(i)$ and $P_Y^{(p)}(j)$ are the marginal histograms. The constant $\varepsilon = 10^{-6}$ ensures numerical stability.

The final local mutual information loss is defined as the negative mean MI across all $N_\text{patch}$ patches, which is fully differentiable and compatible with deep learning frameworks. 
\begin{equation}
\mathcal{L}{\text{MI}} = 1 - \frac{1}{N_\text{patch}} \sum_{p=1}^{N_\text{patch}} \text{MI}^{(p)}(X, Y)
\end{equation}

The segmentation network is supervised using Dice loss on the labeled images, while pseudo-labels generated for the unlabeled images extend supervision across all modalities. We used the BRBS method mainly to achieve semi-supervised liver segmentation. The multi-modality MRI alignment results are generated by applying ANTs \cite{tustison2021antsx} rigid registration, which are found to be better than the BRBS deformable registration results for this task. 

The aligned multi-parametric MRIs and their corresponding liver masks are then used in the classification model for patch extraction from the liver region. 

\subsection{Patch-based Liver Fibrosis Staging}
\label{ssec:cla}
A patch-based classification strategy was employed, leveraging the pathological manifestations of liver fibrosis across different stages. Based on the aligned multi-parametric MRIs, overlapping and multi-channel patches are extracted from the liver region. Several patch sizes (8$\times$8, 16$\times$16, and 32$\times$32) were evaluated, with 16$\times$16 achieving the best performance given the input image size. The multi-channel patches for non-contrast data are three channels (i.e. T1, T2, DWI), and for contrast-enhanced data are seven channels (i.e. T1, T2, DWI, GED1-GED4). To handle missing modalities, zero-filled channels were used.

We hypothesize that liver fibrosis occurs in a local region rather than necessarily affecting the whole liver region, especially for the intermediate stages 2 and 3. Hence, for model training, we only sample patches from Stage~1 and Stage~4 subjects to form the ``good'' and ``bad'' patches, respectively, for a binary classification. The Stage~2 and Stage~3 subjects are excluded from the training and used only for validation purposes. A simplified ResNet-18 backbone was employed for patch-level feature extraction, followed by an MLP classifier. 

During model inference, overlapping patches are extracted from the whole liver region of an unseen subject, and the patch-level classification is performed. The subject-level fibrosis stage is estimated by the percentage of Stage 4 patches over all patches. Thresholds $\tau_{1}$ and $\tau_{2}$ are determined based on a validation set and applied to the percentage value for classifying Stage 1 vs. Stage 2–4 and Stage 4 vs. Stage 1–3, respectively. This percentage value is then converted to a likelihood value between 0 to 1 using piecewise mapping functions based on the thresholds of $\tau_{1}$ and $\tau_{2}$. The resulting probabilities are denoted as $\hat{y}_1$ for substantial fibrosis detection (Stage 1 vs. Stage 2–4) and $\hat{y}_4$ for cirrhosis detection (Stage 4 vs. Stage 1–3). Here is the pipeline of the classification process:

Given a subject with modalities $\{M_i\}_{i=1}^K$ and segmentation mask $\Omega$:  

\begin{enumerate}
    \item \textbf{Patch extraction:}
    \[
    \mathcal{P} = \Big\{\, p_{j}(M_i \odot \Omega) \;\Big|\; i=1,\dots,K, \, j=1,\dots,N \,\Big\}
    \]
     $\mathcal{P}$ contains $N$ overlapping $16\times16$ patches from all $K$ modalities, with zero-filling to handle missing channels.

    \item \textbf{Patch-level classification:}
    \[
    z_j = f_{\text{ResNet18+MLP}}(p_j), \quad z_j \in \{0,1\}
    \]
    where $z_j=1$ denotes a Stage~4-like patch.

    \item \textbf{Subject-level scoring:}
    \[
    s = \frac{1}{N} \sum_{j=1}^N z_j
    \]
    $s$ is the proportion of Stage~4-like patches.

    \item \textbf{Fibrosis probability mapping:}
    \[
    \begin{minipage}{0.48\textwidth}
    \[
    \hat{y}_1 =
    \begin{cases}
     1 - \frac{0.5}{\tau_1}\,s, & 0 \leq s \leq \tau_1 \\[6pt]
     0.5 - \frac{0.5}{1-\tau_1}\,(s-\tau_1), & \tau_1 < s \leq 1
    \end{cases}
    \]
    \end{minipage}
    \hfill
    \hspace{0.5cm} 
    \begin{minipage}{0.48\textwidth}
    \[
    \hat{y}_4 =
    \begin{cases}
    \frac{0.5s}{\tau_2}, & 0 \leq  s \leq \tau_2 \\[6pt]
    \frac{0.5(s-\tau_2)}{(1-\tau_2)} + 0.5, &  \tau_2 < s \leq 1
    \end{cases}
    \]
    \end{minipage}
    \]
    with threshold $\tau_1$  and $\tau_2$ optimized on the validation set.
\end{enumerate}

\section{Experiments and Results}
\label{sec:results}

\subsection{Dataset}
\label{ssec:dataset}

The experiments were conducted on the \textbf{CARE Liver 2025 Track 4} dataset~\cite{liu2025merit}, which contains multi-parametric MRI scans from 610 patients diagnosed with liver fibrosis. Each subject may include part or all of the modalities: T1-weighted (T1), T2-weighted (T2), diffusion-weighted imaging (DWI), and Gd-EOB-DTPA-enhanced dynamic phases (GED1–GED4). Some modalities are missing for certain subjects; however, GED4 is consistently available for all subjects. As summarized in Table~\ref{tab:dataset}, the dataset is collected from multiple centers and vendors, forming three in-distribution (ID) groups: A, B1, and B2.

The training set contains 360 cases, with each group contributing 10 cases with manually annotated liver masks and the remaining cases left unlabeled. The validation set comprises 60 cases (20 per group). The test set includes 190 cases, with 60 from each ID group and 70 out-of-distribution (OOD) cases from vendor C, center C, which are used to assess model generalization.

\begin{table}[t]
    \centering
    \caption{Summary of dataset splits in CARE Liver 2025 Track 4. Values are shown as total cases (labeled liver mask cases).}
    \setlength{\tabcolsep}{10pt} 
    \renewcommand{\arraystretch}{1.2} 
    \begin{tabular}{lccccc}
        \hline
        \textbf{Dataset} & \textbf{A} & \textbf{B1} & \textbf{B2} & \textbf{OOD (C)} & \textbf{Total} \\
        \hline
        Training         & 130 (10)   & 170 (10)    & 60 (10)     & --              & 360 (30) \\
        Validation       & 20 (20)    & 20 (20)     & 20 (20)     & --              & 60 (60) \\
        Test             & 60 (60)    & 60 (60)     & 60 (60)     & 70 (70)         & 250 (250) \\
        \hline
    \end{tabular}
    \label{tab:dataset}
\end{table}


For the segmentation experiments, we used 62 images (31 from T1 and 31 from T2) as a validation set, while all remaining images (566) were included in the training set. 

For training the classification model, only 87 Stage 1 and 150 Stage 4 subjects were used. To address the substantial class imbalance, geometric transformations, such as flipping and rotation, were applied to both liver masks and all corresponding modality channels. This augmentation strategy balanced the number of patches between classes and mitigated the potential bias in model training. The validation phase included subjects from all four stages, comprising 10 Stage 1, 32 Stage 2, 64 Stage 3, and 17 Stage 4 subjects. 


\subsection{Parameter Setting}
\label{ssec:setting}
All segmentation experiments were conducted using a learning rate of $1 \times 10^{-4}$, with training performed for 500 epochs, each epoch consisting of 200 iterations. The network was trained from scratch using all available data (both labeled and unlabeled), as described in Section~\ref{ssec:dataset}. 

In the classification model, a learning rate of $1 \times 10^{-4}$ was employed. Given the large number of patches and their high similarity, training was limited to 10 epochs. All models were trained from scratch using the cross-entropy loss. Dynamic learning rate scheduling was applied at the iteration level, with the learning rate adjusted every 5000 iterations. We employed 4-fold cross-validation to determine the thresholds $\tau_1$ and $\tau_2$, which were set to 0.37 and 0.66 for non-contrast MRIs and 0.35 and 0.70 for contrast MRIs. 

\subsection{Liver Segmentation Results}
In the challenge of liver segmentation subtasks (LiSeg), our method was evaluated on all MRI modalities of both ID and OOD tasks. 

\label{ssec:seg-results}

\vspace{-2pt} 
\setlength{\textfloatsep}{8pt plus 1pt minus 2pt}
\setlength{\floatsep}{8pt plus 1pt minus 2pt}
\begin{center}
\captionof{table}{Segmentation results on the CARE Liver 2025 validation set.}
\footnotesize
\renewcommand{\arraystretch}{1.15}
\setlength{\tabcolsep}{6.5pt}
\begin{tabular}{lcccccccc}
  \toprule
  \textbf{Method} &
    \multicolumn{2}{c}{\textbf{GED4}} &
    \multicolumn{2}{c}{\textbf{T1}} &
    \multicolumn{2}{c}{\textbf{T2}} &
    \multicolumn{2}{c}{\textbf{DWI}} \\
  & Dice↑ & HD↓ & Dice↑ & HD↓ & Dice↑ & HD↓ & Dice↑ & HD↓ \\
  \midrule
  BRBS-NCC & 0.9047 & 42.12 & 0.8935 & 73.90 & 0.7423 & 52.30 & 0.6521 & 45.67 \\
  BRBS-MI  & 0.9518 & 36.81 & 0.9345 & 66.75 & 0.8323 & 35.95 & 0.8323 & 35.95 \\
  \bottomrule
\label{tab:seg-results-all}
\end{tabular}
\end{center}
\vspace{-14pt} 

\vspace{-2pt} 
\setlength{\textfloatsep}{8pt plus 1pt minus 2pt}
\setlength{\floatsep}{8pt plus 1pt minus 2pt}
\begin{table}[!htbp]
\centering
\caption{Comparison of leaderboard results on the test set for Non-Contrast subtasks.
The In-Distribution data (vendors A, B1, and B2) and Out-of-Distribution data (vendor C) 
are evaluated separately. Only the top five teams are shown.}
\label{tab:seg_results}



\par\noindent
\begin{minipage}[t]{0.45\linewidth}
\centering
\begin{tabular}{lcc}
\toprule
Team & Dice(\%)  & HD(mm) \\
\midrule
BIGS2 & 94.34 & 38.06 \\
\textbf{Sigma(Ours)} & \textbf{93.18} & \textbf{58.99} \\
CitySJTU & 92.62 & 27.80 \\
BioDreamer & 91.96 & 31.83 \\
MIHL & 86.42 & 163.91 \\
\bottomrule
\end{tabular}
\caption*{(a) T1 (In-Distribution)}
\end{minipage}
\hfill
\begin{minipage}[t]{0.45\linewidth}
\centering
\begin{tabular}{lcc}
\toprule
Team & Dice(\%) & HD(mm) \\
\midrule
BIGS2 & 95.48 & 22.11 \\ 
\textbf{Sigma(Ours)} & \textbf{95.03} & \textbf{29.04}\\
CitySJTU & 94.44 & 25.54  \\
BioDreamer & 94.18 & 28.24 \\
MIHL & 73.51 & 102.79  \\
\bottomrule
\end{tabular}
\caption*{(b) T1 (Out-Of-Distribution)}
\end{minipage}

\vspace{8pt}

\par\noindent
\begin{minipage}[t]{0.45\linewidth}
\centering
\begin{tabular}{lcc}
\toprule
Team & Dice(\%)  & HD(mm) \\
\midrule
CitySJTU & 88.90 & 34.60  \\
BIGS2 & 87.22& 35.24\\
BioDreamer & 82.01 & 37.16 \\
\textbf{Sigma(Ours)} & \textbf{75.68} & \textbf{75.44} \\
MIHL & 20.24 & 121.48 \\
\bottomrule
\end{tabular}
\caption*{(c) T2 (In-Distribution)}
\end{minipage}
\hfill
\begin{minipage}[t]{0.45\linewidth}
\centering
\begin{tabular}{lcc}
\toprule
Team & Dice(\%) & HD(mm) \\
\midrule
BIGS2 & 90.15& 35.17 \\ 
CitySJTU & 88.94 & 24.52 \\
BioDreamer & 84.57 & 25.84 \\
\textbf{Sigma(Ours)} & \textbf{62.08} & \textbf{72.29} \\
MIHL & 7.14 & 118.06  \\
\bottomrule
\end{tabular}
\caption*{(d) T2 (Out-Of-Distribution)}
\end{minipage}

\vspace{8pt}

\par\noindent
\begin{minipage}[t]{0.45\linewidth}
\centering
\begin{tabular}{lcc}
\toprule
Team & Dice(\%)  & HD(mm) \\
\midrule
CitySJTU & 83.74 & 26.96 \\
\textbf{Sigma(Ours)} & \textbf{82.34} & \textbf{34.19} \\
BIGS2 & 80.29& 31.04 \\
BioDreamer & 72.10 & 31.33 \\
MIHL & 26.12 & 99.25\\
\bottomrule
\end{tabular}
\caption*{(e) DWI (In-Distribution)}
\end{minipage}
\hfill
\begin{minipage}[t]{0.45\linewidth}
\centering
\begin{tabular}{lcc}
\toprule
Team & Dice(\%) & HD(mm) \\
\midrule
\textbf{Sigma(Ours)} & \textbf{90.41} & \textbf{19.72} \\ 
CitySJTU & 88.57 & 11.21 \\
BIGS2 & 84.81 & 11.21 \\
BioDreamer & 78.12 & 15.96 \\
MIHL & 45.67 & 51.76\\
\bottomrule
\end{tabular}
\caption*{(f) DWI (Out-Of-Distribution)}
\end{minipage}

\end{table}

We first evaluated our method on the validation set for refining our model. We compared two configurations of the BRBS framework:  
(1) \textbf{BRBS-NCC}, the original BRBS using NCC loss, trained with 30 annotated GED4 images and all unlabeled in the CARE Liver 2025 dataset;  
(2) \textbf{BRBS-MI}, where NCC is replaced by the MI loss, using the same training data as BRBS-NCC.

As shown in Table~\ref{tab:seg-results-all}, replacing NCC by MI led to consistent improvement across all MRI sequences. The MI-based model (BRBS-MI) achieved higher Dice scores (Dice) and lower Hausdorff Distances (HD), especially on T2 and DWI, where modality differences were more pronounced. This demonstrates that MI better handles cross-modality variations during registration and improves label propagation.


Our method was further evaluated and compared to other teams using the test set, as shown in Table \ref{tab:seg_results}. Our method produced Dice values of 93.18\% and 95.03\% for T1 ID and T1 OOD respectively, and ranked the second place. 
The DWI results were particularly noteworthy: our model achieved the second best ID segmentation performance (Dice = 82.34\%), and the best OOD segmentation performance (Dice = 90.41\%), confirming its robustness to domain shifts and signal variations. 
However, segmentation on the T2 modality remained a relative lower performance compared to other methods, with Dice scores of 75.68\% (ID) and 62.08\% (OOD), which ranked us the fourth place. This decreased performance on T2 MRI, could be caused by the higher intensity heterogeneity and its limited representation in the training data.

\subsection{Liver Fibrosis Classification Results}
\label{ssec:cls-results}

\begin{table}[htbp]
\centering
\caption{Comparison of leaderboard results on the test set for liver fibrosis staging tasks.
The In-Distribution data (vendors A, B1, and B2) and Out-of-Distribution data (vendor C) 
are evaluated separately. Only the top five teams are shown.}
\label{tab:FS_results}

\begin{minipage}[t]{0.45\linewidth}
\centering
\begin{tabular}{lcc}
\toprule
Team & ACC(\%)  & AUC(\%) \\
\midrule
\textbf{Sigma(Ours)} & \textbf{75.83} & \textbf{83.92} \\
WSQ & 72.50 & 78.31 \\
Team space & 71.67 & 79.39 \\
CitySJTU & 70.83 & 75.10 \\
BioDreamer & 68.33 & 77.19 \\
\bottomrule
\end{tabular}
\caption*{(a) Contrast-Enhanced in S4 vs. S1-S3 (In-Distribution) }
\end{minipage}
\hfill
\begin{minipage}[t]{0.45\linewidth}
\centering
\begin{tabular}{lcc}
\toprule
Team & ACC(\%)  & AUC(\%)\\
\midrule
WSQ & 80.83 & 84.50 \\
Team space & 79.17 & 80.24 \\
\textbf{Sigma(Ours)} & \textbf{77.50} & \textbf{83.42} \\
CitySJTU & 75.00 & 78.45 \\
BioDreamer & 74.58 & 78.11 \\
\bottomrule
\end{tabular}
\caption*{(b) Contrast-Enhanced in S1 vs. S2-S4 (In-Distribution) }
\end{minipage}

\vspace{8pt}

\par\noindent
\begin{minipage}[t]{0.45\linewidth}
\centering
\begin{tabular}{lcc}
\toprule
Team & ACC(\%)  & AUC(\%) \\
\midrule
BioDreamer & 63.71 & 67.40 \\
\textbf{Sigma(Ours)} & \textbf{58.57} & \textbf{54.12} \\
Team space & 52.86 & 60.22 \\
CitySJTU & 41.43 & 46.81 \\
WSQ & 41.43 & 66.95 \\
\bottomrule
\end{tabular}
\caption*{(c) Contrast-Enhanced in S4 vs. S1-S3 (Out-Of-Distribution)}
\end{minipage}
\hfill
\begin{minipage}[t]{0.45\linewidth}
\centering
\begin{tabular}{lcc}
\toprule
Team & ACC(\%)  & AUC(\%) \\
\midrule
\textbf{Sigma(Ours)} & \textbf{92.86} & \textbf{75.38} \\ 
NW-Radio & 92.86 & 52.62\\
BioDreamer & 92.86 & 47.68 \\
WSQ & 82.86 & 37.23  \\
Team space & 64.29 & 64.62  \\
\bottomrule
\end{tabular}
\caption*{(d) Contrast-Enhancedin S1 vs. S2-S4 (Out-Of-Distribution)}
\end{minipage}

\vspace{8pt}

\par\noindent
\begin{minipage}[t]{0.45\linewidth}
\centering
\begin{tabular}{lcc}
\toprule
Team & ACC(\%) & AUC(\%) \\
\midrule
Team space & 74.17 & 81.33 \\
potato & 72.50 & 77.22 \\
CitySJTU & 71.67 & 78.61 \\
BioDreamer & 70.83 & 77.89 \\
\textbf{Sigma(Ours)} & \textbf{70.00} & \textbf{77.22} \\
\bottomrule
\end{tabular}
\caption*{(e) Non-Contrast in S4 vs. S1-S3 (In-Distribution)}
\end{minipage}
\hfill
\begin{minipage}[t]{0.45\linewidth}
\centering
\begin{tabular}{lcc}
\toprule
Team & ACC(\%) & AUC(\%) \\
\midrule
Team space & 76.67 & 83.93 \\ 
TeamZhang & 74.17 & 68.48\\
potatp & 71.67 & 78.51 \\
BioDreamer  & 70.00 & 77.11  \\
CitySJTU & 70.00 & 73.71  \\
\bottomrule
\end{tabular}
\caption*{(f) Non-Contrast in S1 vs. S2-S4 (In-Distribution)}
\end{minipage}

\vspace{8pt}

\par\noindent
\begin{minipage}[t]{0.45\linewidth}
\centering
\begin{tabular}{lcc}
\toprule
Team &  ACC(\%) & AUC(\%) \\
\midrule
\textbf{Sigma(Ours)} & \textbf{71.43} & \textbf{69.47} \\
potato & 70.00 & 71.51 \\
TeamZhang & 64.29 & 68.83 \\
Team space & 64.29 & 52.73 \\
CitySJTU & 42.86 & 49.31 \\
\bottomrule
\end{tabular}
\caption*{(g) Non-Contrast in S4 vs. S1-S3 (Out-Of-Distribution)}
\end{minipage}
\hfill
\begin{minipage}[t]{0.45\linewidth}
\centering
\begin{tabular}{lcc}
\toprule
Team & ACC(\%)  & AUC(\%) \\
\midrule
\textbf{Sigma(Ours)} & \textbf{92.86} & \textbf{86.31} \\ 
NW-Radio & 92.86 & 40.31\\
TeamZhang & 91.43 & 71.38 \\
Team space & 88.57 & 61.23  \\
BioDreamer & 88.29 & 59.40  \\
\bottomrule
\end{tabular}
\caption*{(h) Non-Contrast in S1 vs. S2-S4 (Out-Of-Distribution)}
\end{minipage}

\end{table}

Based on the segmentation results, we further evaluated the classification performance of liver fibrosis staging task, and compared to other teams. Task 1 refers to the classification of Stage 4 vs. Stage 1–3, and Task 2 refers to the classification of Stage 1 vs. Stage 2–4. The area under the ROC (AUC) and classification accuracy (ACC) were calculated based on the results obtained by running our docker file on the challenge organizer's server. Table \ref{tab:FS_results} presents the top five teams’ performance across all fibrosis staging subtasks on the test set.


For the contrast-enhanced modality results (Table \ref{tab:FS_results} (a)-(d)), our method achieved the best performance in ID cases (ACC = 75.83\%) for Task 1, and ranked the third for Task 2 with an ACC of 77.50\%. For the OOD cases, our method achieved the second best (ACC=58.57\%) and the best performance (ACC=92.86\%) on Task 1 and Task 2 respectively. The results show that the model achieved high accuracies in both ID and OOD cases, demonstrating its ability to correctly stage liver fibrosis in both tasks.

For the non-contrast modality results (Table \ref{tab:FS_results} (e)-(h)), our method achieved relatively lower performance than other methods for the ID cases (ranked the 5th). However, it achieved the best performance for OOD cases on both Task 1 (ACC=71.43\%) and Task 2 (ACC=92.86\%). This indicates that our model generalized well on test cases of unseen centers.

\begin{figure*}[h]
    \centering
    \includegraphics[width=0.9\linewidth]{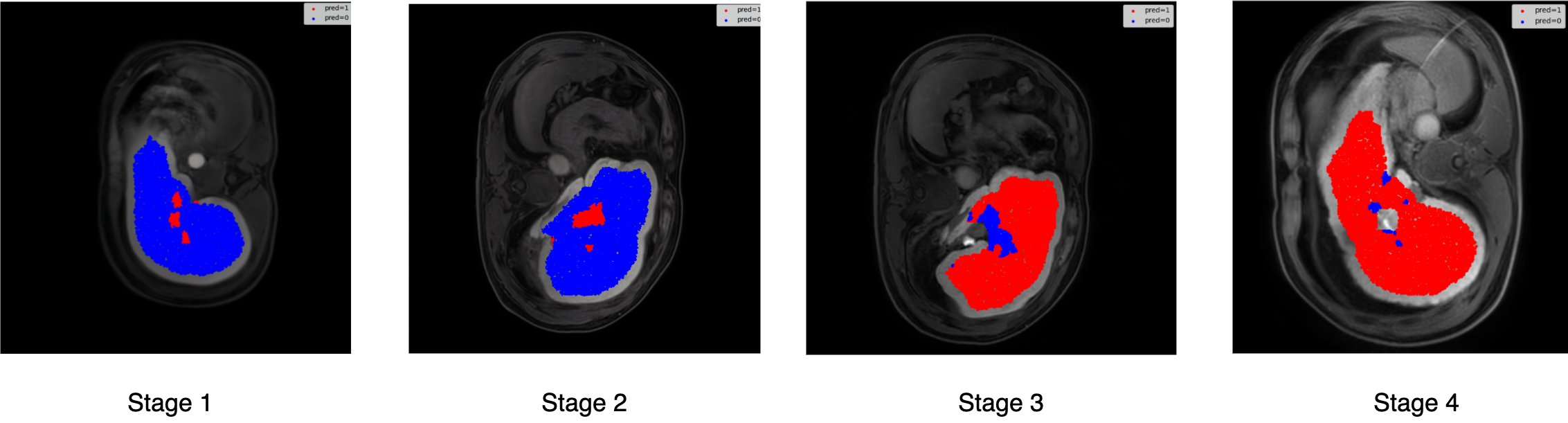}
    \caption{Visualization of patch-level predictions at different fibrosis stages (Stage 1–4). Blue represents stage 1 patches (pred = 0), while red represents stage 4 patches (pred = 1). The percentage of blue and red patches varies with disease stage, reflecting the progression from early stages to advanced stages.}
    \label{fig:care2025-visual}
\end{figure*}

One advantage of using our patch-based method is that the percentage of Stage-4 patches can be visualized for certain interpretability. Figure \ref{fig:care2025-visual} shows the visualization of the patch-level prediction for one example subject in each stage. Red presents the stage 4 patch, and blue refers to the stage 1 patch. It can be seen that as the fibrosis severity increases, the percentage of stage-4 patches increases as well. 

Overall, the liver fibrosis staging results reveal that our method achieves top-tier performance in both contrast-enhanced and non-contrast modalities, with strong OOD generalization. The classification accuracy could be dependent on the liver segmentation performance (e.g. particularly lower for T2 MRI), as inaccurate liver region of interest could affect the precision of patch sampling. 



\section{Conclusions}
\label{sec:conclusions}

We proposed a fully automated framework for liver fibrosis staging that integrates semi-supervised segmentation, multi-modality registration and a patch-based classifier. The method achieves robust segmentation across GED4, T1, T2, and DWI MRIs, and demonstrates strong performance in liver fibrosis staging, particularly using three-channel non-contrast MRIs. Importantly, the patch-based design enables visualization of localized pathological regions, offering clinically meaningful insights for objective liver fibrosis assessment. Future work will focus on streamlining the framework to improve the efficiency and accuracy. 

%
%
%
\clearpage
\bibliographystyle{splncs04}
\bibliography{refs}
%




\end{document}